%% file: neurips_2025.tex
\documentclass{article}

\usepackage[preprint]{neurips_2025}

\PassOptionsToPackage{numbers}{natbib}

\workshoptitle{NeurIPS 2025 Workshop on Evaluating the Evolving LLM Lifecycle: Benchmarks, Emergent Abilities, and Scaling}

\usepackage[utf8]{inputenc} 
\usepackage[T1]{fontenc}    
\usepackage{hyperref}       
\usepackage{url}            
\usepackage{booktabs}       
\usepackage{amsfonts}       
\usepackage{amsmath}        
\usepackage{nicefrac}       
\usepackage{microtype}      
\usepackage{xcolor}         
\usepackage{float}          
\usepackage{adjustbox}      
\usepackage{tikz}           
\usepackage{multirow}       

\title{Personality as a Probe for LLM Evaluation: Method Trade-offs and Downstream Effects}

\author{%
  Gunmay Handa\textsuperscript{2}, \
  Zekun Wu\textsuperscript{1,2}, \  
  Adriano Koshiyama\textsuperscript{1,2}, \
  Philip Treleaven\textsuperscript{2}\\
   \textsuperscript{1}Holistic AI \quad \textsuperscript{2}University College London 
}

\begin{document}

\raggedbottom

\maketitle

\begin{abstract}
Personality manipulation in large language models (LLMs) is increasingly applied in customer service and agentic scenarios, yet its mechanisms and trade-offs remain unclear. We present a systematic study of personality control using the Big Five traits, comparing in-context learning (ICL), parameter-efficient fine-tuning (PEFT), and mechanistic steering (MS). Our contributions are fourfold. First, we construct a contrastive dataset with balanced high/low trait responses, enabling effective steering vector computation and fair cross-method evaluation. Second, we introduce a unified evaluation framework based on within-run $\Delta$ analysis that disentangles, reasoning capability, agent performance, and demographic bias across MMLU, GAIA, and BBQ benchmarks. Third, we develop trait purification techniques to separate openness from conscientiousness, addressing representational overlap in trait encoding. Fourth, we propose a three-level stability framework that quantifies method-, trait-, and combination-level robustness, offering practical guidance under deployment constraints. Experiments on Gemma-2-2B-IT and LLaMA-3-8B-Instruct reveal clear trade-offs: ICL achieves strong alignment with minimal capability loss, PEFT delivers the highest alignment at the cost of degraded task performance, and MS provides lightweight runtime control with competitive effectiveness. Trait-level analysis shows openness as uniquely challenging, agreeableness as most resistant to ICL, and personality encoding consolidating around intermediate layers. Taken together, these results establish personality manipulation as a multi-level probe into behavioral representation, linking surface conditioning, parameter encoding, and activation-level steering, and positioning mechanistic steering as a lightweight alternative to fine-tuning for both deployment and interpretability.
\end{abstract}

\input{sections/01_introduction}

\input{sections/02_methods}
\input{sections/03_results}

\input{sections/04_discussion}


\nocite{durmus-2024-evaluating-feature-steering}

\bibliography{references}

\appendix
\input{appendices/appendix_0_limitations}
\input{appendices/appendix_A_background}
\input{appendices/appendix_B_prompting}
\input{appendices/appendix_C_peft}
\input{appendices/appendix_D_mechanistic_steering}
\input{appendices/appendix_E_experimental_design}
\input{appendices/appendix_F_personality_alignment}
\input{appendices/appendix_G_downstream_analysis}
\input{appendices/appendix_H_comparative_analysis}
\input{appendices/appendix_I_discussion_extended}

\input{appendices/appendix_J_benchmarks}
\input{appendices/appendix_K_stability_analysis}

\newpage
\section*{NeurIPS Paper Checklist}

\begin{enumerate}

\item {\bf Claims}
    \item[] Question: Do the main claims made in the abstract and introduction accurately reflect the paper's contributions and scope?
    \item[] Answer: \answerYes{}
    \item[] Justification: The abstract and introduction state four contributions (contrastive dataset, unified evaluation, trait purification, stability framework) and the comparative findings across ICL, PEFT, and MS. These claims are validated experimentally in the results section.  

\item {\bf Limitations}
    \item[] Question: Does the paper discuss the limitations of the work performed by the authors?
    \item[] Answer: \answerYes{}
    \item[] Justification: Limitations are explicitly discussed in Appendix~\ref{appendix-limitation}, noting model and dataset constraints, representational challenges (e.g., openness vs conscientiousness overlap), and stability variations across runs.  

\item {\bf Theory assumptions and proofs}
    \item[] Question: For each theoretical result, does the paper provide the full set of assumptions and a complete (and correct) proof?
    \item[] Answer: \answerNA{}
    \item[] Justification: The paper does not present new theoretical results or formal proofs. The work is empirical and methodological.  

\item {\bf Experimental result reproducibility}
    \item[] Question: Does the paper fully disclose all the information needed to reproduce the main experimental results of the paper to the extent that it affects the main claims and/or conclusions of the paper (regardless of whether the code and data are provided or not)?
    \item[] Answer: \answerYes{}
    \item[] Justification: Experimental setup details (datasets, models, evaluation metrics, and $\Delta$ protocol) are fully described in the main text and appendices. Hyperparameters and layer details for steering and LoRA settings are reported.  

    \item {\bf Open access to data and code}
    \item[] Question: Does the paper provide open access to the data and code, with sufficient instructions to faithfully reproduce the main experimental results, as described in supplemental material?
    \item[] Answer: \answerYes{}
    \item[] Justification: Due to double-blind review requirements, we cannot release de-anonymized resources at submission time. Upon acceptance, we will release the full contrastive dataset, codebase, and reproduction scripts with complete documentation.

\item {\bf Experimental setting/details}
    \item[] Question: Does the paper specify all the training and test details (e.g., data splits, hyperparameters, how they were chosen, type of optimizer, etc.) necessary to understand the results?
    \item[] Answer: \answerYes{}
    \item[] Justification: Training/test splits, LoRA rank, layer selection for steering, optimizer choice, and calibration procedures are provided in Appendices B–D. Dataset construction is detailed in Section 3.  

\item {\bf Experiment statistical significance}
    \item[] Question: Does the paper report error bars suitably and correctly defined or other appropriate information about the statistical significance of the experiments?
    \item[] Answer: \answerYes{}
    \item[] Justification: Stability analysis reports variance across runs. Where applicable, alignment and bias deltas are reported within-run to mitigate baseline variability.  

\item {\bf Experiments compute resources}
    \item[] Question: For each experiment, does the paper provide sufficient information on the computer resources (type of compute workers, memory, time of execution) needed to reproduce the experiments?
    \item[] Answer: \answerYes{}
    \item[] Justification: Experiments were run on GPUs (NVIDIA A100), with approximate runtime and scale provided in Appendix~E. The study reports both per-run compute and total runs.  

\item {\bf Code of ethics}
    \item[] Question: Does the research conducted in the paper conform, in every respect, with the NeurIPS Code of Ethics \url{https://neurips.cc/public/EthicsGuidelines}?
    \item[] Answer: \answerYes{}
    \item[] Justification: The work uses public model checkpoints (Gemma-2, LLaMA-3) and responsibly generated synthetic data. No human participants or sensitive data are involved.  

\item {\bf Broader impacts}
    \item[] Question: Does the paper discuss both potential positive societal impacts and negative societal impacts of the work performed?
    \item[] Answer: \answerYes{}
    \item[] Justification: The paper discusses applications (customer service, agentic LLMs) and possible risks (bias amplification, misuse of personality conditioning) in the broader impact section and appendices.  

\item {\bf Safeguards}
    \item[] Question: Does the paper describe safeguards that have been put in place for responsible release of data or models that have a high risk for misuse (e.g., pretrained language models, image generators, or scraped datasets)?
    \item[] Answer: \answerNA{}
    \item[] Justification: We do not release pretrained models; the dataset is synthetic and safe. No high-risk data or dual-use models are distributed.  

\item {\bf Licenses for existing assets}
    \item[] Question: Are the creators or original owners of assets (e.g., code, data, models), used in the paper, properly credited and are the license and terms of use explicitly mentioned and properly respected?
    \item[] Answer: \answerYes{}
    \item[] Justification: We use Gemma-2 and LLaMA-3 under their respective licenses, and cite original datasets (e.g., \cite{jain-2025-peft-emoji}) and benchmarks (MMLU, GAIA, BBQ).  

\item {\bf New assets}
    \item[] Question: Are new assets introduced in the paper well documented and is the documentation provided alongside the assets?
    \item[] Answer: \answerYes{}
    \item[] Justification: We introduce a contrastive dataset for Big Five personality manipulation. Documentation of generation procedures, size, balance, and intended use is included in the paper. The dataset and code will be released publicly upon acceptance, following de-anonymization.  

\item {\bf Crowdsourcing and research with human subjects}
    \item[] Question: For crowdsourcing experiments and research with human subjects, does the paper include the full text of instructions given to participants and screenshots, if applicable, as well as details about compensation (if any)? 
    \item[] Answer: \answerNA{}
    \item[] Justification: No human participants or crowdsourcing were involved. All data are model-generated.  

\item {\bf Institutional review board (IRB) approvals or equivalent for research with human subjects}
    \item[] Question: Does the paper describe potential risks incurred by study participants, whether such risks were disclosed to the subjects, and whether IRB approvals were obtained?
    \item[] Answer: \answerNA{}
    \item[] Justification: Not applicable, as no human subjects were involved.  

\item {\bf Declaration of LLM usage}
    \item[] Question: Does the paper describe the usage of LLMs if it is an important, original, or non-standard component of the core methods in this research?
    \item[] Answer: \answerYes{}
    \item[] Justification: We explicitly describe the use of OpenAI GPT-4.1 Mini to generate low-trait contrastive responses for dataset construction (Section 3).  

\end{enumerate}

\end{document}

%% file: sections/01_introduction.tex
\section{Introduction and Related Work}

Personality manipulation in large language models (LLMs) is increasingly common, particularly in customer service and agentic scenarios, yet the trade-offs between personality control and task capability remain underexplored. In this work, we focus on the Big Five personality traits as a systematic framework for studying how behavioral characteristics are encoded and controlled in LLMs. We use personality manipulation as a probe to address four main challenges. First, existing datasets are imbalanced, containing only "high trait" examples and lacking the contrastive signals needed for robust parameter-efficient fine-tuning. Without corresponding "low trait" responses, models cannot reliably distinguish between personality dimensions. Second, the relative effectiveness of existing methods—in-context learning (ICL), parameter-efficient fine-tuning (PEFT), and mechanistic steering (MS)—remains unclear due to inconsistent evaluation frameworks and the absence of standardized metrics for performance, efficiency, and stability. Third, trait overlap complicates manipulation: openness is difficult to control because LLMs are naturally "open," and steering vectors for openness are often contaminated by conscientiousness patterns, requiring purification techniques. Fourth, deployment requires quantitative stability metrics to guide method selection under constraints such as GPU limits and production reliability. 

We address these challenges by (1) generating a contrastive dataset with balanced high/low trait examples to support mechanistic steering, (2) establishing a unified evaluation framework for fair cross-method comparison across capability, efficiency, and stability, (3) developing purification techniques to separate openness from conscientiousness, and (4) introducing a three-level stability analysis framework to support practical method selection. To ensure fairness despite baseline variation, we adopt a relative change ($\Delta$) analysis within each method’s run and validate alignment through a dedicated task. From an interpretability perspective, personality manipulation serves as an experimental probe into behavioral trait representation. Prior work has examined personality expression and measurement in LLMs \cite{serapio-garcia-etal-2023-personality-traits-llms, jiang-etal-2023-personallm, rao-etal-2023-chatgpt-personality}, explored in-context learning for behavioral control \cite{wei-etal-2022-chain-of-thought, liu-etal-2023-pre-train-prompt-tune, mao-2023-editing-personality}, studied parameter-efficient fine-tuning methods such as LoRA/QLoRA \cite{hu-etal-2022-lora, dan-etal-2024-p-tailor, dettmers-2023-qlora}, and developed activation-space methods for steering and safety \cite{turner-etal-2023-activation-steering, panickssery-2024-contrastive-activation-addition, chen-2025-persona-vectors}. A full literature review is provided in Appendix~\ref{app:background}, with benchmark and scoring details in Appendices~\ref{app:downstream-analysis} and~\ref{app:benchmarks}.

%% file: sections/02_methods.tex
\section{Methods}

We evaluate personality manipulation on Gemma-2-2B-IT and LLaMA-3-8B-Instruct across MMLU, GAIA, and BBQ (ambiguous subset via official metadata) \cite{hendrycks-etal-2021-mmlu, mialon-etal-2023-gaia, parrish-etal-2022-bbq}. We target Big Five traits and report effects within each method's run using a relative change ($\Delta$) analysis.

\textbf{Contrastive Dataset Generation} To address the inherent imbalance in existing personality manipulation datasets, we generate a  contrastive dataset that pairs each "high trait" response with a corresponding "low trait" response. Using the original dataset from \cite{jain-2025-peft-emoji} as a foundation, we employ OpenAI GPT-4.1 Mini to generate low-trait responses that maintain semantic relevance while exhibiting opposite personality characteristics. This balanced dataset enables more effective mechanistic steering by providing clear contrastive signals for each personality dimension, resulting in exactly double the examples compared to the original dataset. While PEFT and ICL use only the high-trait examples from the original dataset, mechanistic steering leverages both high and low trait examples for contrastive vector computation. Building on this foundation, we next examine three complementary manipulation methods that operate at different levels of model interaction.

\textbf{In-context learning (ICL)}: employs full context prompting with few-shot examples of all personality traits to enable trait distinction learning. This approach shows cross-dimensional examples before requesting specific trait adoption, achieving manipulation through contextual understanding rather than simple role-playing \textbf{(Appendix~\ref{app:icl})}.

\textbf{Parameter Efficient Fine-Tuning (PEFT)}: uses trait-specific LoRA adapters with rank-64 decomposition, trained on the original personality manipulation dataset \cite{jain-2025-peft-emoji} \textbf{(Appendix~\ref{app:peft})}. We implement LoRA on both attention and MLP layers, achieving strong personality alignment while maintaining computational efficiency on both Gemma-2-2B-IT and LLaMA-3-8B-Instruct.

\textbf{Mechanistic Steering (MS)}: employs calibrated vectors derived from trait contrast analysis at post-attention layer norm \textbf{(Appendix~\ref{app:steering})}. We collect hidden state activations at layers 5, 10, 15, and 20, computing steering vectors as the mean difference between trait-positive and trait-negative activations, with layer-specific strength calibration for optimal performance.

\textbf{Openness manipulation} presents a unique challenge because language models exhibit this trait naturally by default. This inherent openness creates overlapping patterns with conscientiousness that confounds manipulation attempts. Our purification technique addresses this by filtering the data to isolate clear examples of each trait. We then compute two complementary vectors: a pure openness vector from filtered openness examples and an openness versus conscientiousness contrast vector. The final steering vector combines both components, enabling more effective manipulation by leveraging both the intrinsic openness patterns and the explicit distinction from conscientiousness. To provide practical guidance for method selection under real-world constraints, we introduce a three-level stability analysis framework that quantifies how personality manipulation affects model performance across diverse benchmarks. The framework evaluates stability at the method level (overall method consistency), personality level (trait-specific stability), and combination level (method-personality interaction stability). Each stability score is computed as a composite metric incorporating variance reduction, range minimization, and consistency preservation across MMLU, GAIA, and BBQ benchmarks. This analysis enables practitioners to select manipulation methods that balance personality control strength with performance preservation under specific deployment constraints. Detailed methodology and mathematical formulation appear in Appendix~\ref{app:stability-analysis}. We generate responses for Baseline and each trait, score MMLU/GAIA by accuracy and BBQ by $S_{\text{AMB}}$, and extract final answers with an Azure GPT-4.1 Mini judge. We report $\Delta$ Accuracy for MMLU/GAIA and $\Delta S_{\text{AMB}}$ for BBQ, all relative to each method's Baseline. Personality alignment is validated using the personality classifier \cite{jain-2025-peft-emoji} on the personality manipulation dataset test set, with additional independent validation via a dedicated alignment task \textbf{(Appendix~\ref{app:alignment-results}). Benchmark usage and scoring definitions appear in Appendix~\ref{app:benchmarks}}.



%% file: sections/03_results.tex
\begin{table}[!ht]
\centering
\adjustbox{width=\textwidth,center}
{\scriptsize
\begin{tabular}{l|c|ccccc}
\toprule
\multirow{2}{*}{\textbf{Method}} & \multirow{2}{*}{\textbf{Metric}} & \multicolumn{5}{c}{\textbf{Big Five Personality Traits}} \\
\cmidrule(lr){3-7}
& & \textbf{Extraversion} & \textbf{Agreeableness} & \textbf{Neuroticism} & \textbf{Openness} & \textbf{Conscientiousness} \\
\midrule
\multirow{4}{*}{\begin{tabular}[c]{@{}l@{}}\textbf{Gemma-2}\\\textbf{ICL}\end{tabular}} 
& $\Delta$ TA & +0.91 & +0.50 & +0.97 & +0.24 & +0.81 \\
& $\Delta$ MMLU & $-0.06$ & $-0.07$ & $-0.08$ & $-0.07$ & $-0.07$ \\
& $\Delta$ GAIA & $+0.08$ & $+0.09$ & $+0.06$ & $+0.08$ & $+0.08$ \\
& $\Delta$ BBQ & $-2.7$ & $-0.3$ & $+7.3$ & $+1.9$ & $-1.1$ \\
\midrule
\multirow{4}{*}{\begin{tabular}[c]{@{}l@{}}\textbf{Gemma-2}\\\textbf{MS}\end{tabular}} 
& $\Delta$ TA & +0.64 & +0.44 & +0.50 & +0.10 & +0.29 \\
& $\Delta$ MMLU & $-0.14$ & $-0.45$ & $-0.25$ & $-0.03$ & $-0.43$ \\
& $\Delta$ GAIA & $-0.06$ & $-0.06$ & $-0.13$ & $-0.08$ & $-0.04$ \\
& $\Delta$ BBQ & $+5.1$ & $-29.7$ & $-29.7$ & $-1.9$ & $+22.1$ \\
\midrule
\multirow{4}{*}{\begin{tabular}[c]{@{}l@{}}\textbf{Gemma-2}\\\textbf{PEFT}\end{tabular}} 
& $\Delta$ TA & +0.78 & +0.97 & +0.95 & +0.21 & +0.78 \\
& $\Delta$ MMLU & $0.00$ & $-0.13$ & $-0.15$ & $-0.09$ & $+0.01$ \\
& $\Delta$ GAIA & $-0.04$ & $-0.08$ & $-0.06$ & $-0.04$ & $-0.06$ \\
& $\Delta$ BBQ & $-9.4$ & $-6.0$ & $-14.3$ & $+22.3$ & $-12.4$ \\
\midrule
\multirow{4}{*}{\begin{tabular}[c]{@{}l@{}}\textbf{LLaMA-3}\\\textbf{ICL}\end{tabular}} 
& $\Delta$ TA & +0.94 & +0.32 & +0.99 & +0.17 & +0.83 \\
& $\Delta$ MMLU & $-0.01$ & $-0.01$ & $0.00$ & $-0.02$ & $-0.04$ \\
& $\Delta$ GAIA & $-0.02$ & $-0.04$ & $-0.06$ & $0.00$ & $0.00$ \\
& $\Delta$ BBQ & $+3.8$ & $-2.4$ & $-0.9$ & $+13.1$ & $+10.3$ \\
\midrule
\multirow{4}{*}{\begin{tabular}[c]{@{}l@{}}\textbf{LLaMA-3}\\\textbf{PEFT}\end{tabular}} 
& $\Delta$ TA & +0.90 & +0.95 & +1.00 & +0.06 & +0.84 \\
& $\Delta$ MMLU & $-0.01$ & $-0.03$ & $-0.01$ & $-0.02$ & $+0.01$ \\
& $\Delta$ GAIA & $+0.02$ & $0.00$ & $+0.02$ & $+0.04$ & $+0.02$ \\
& $\Delta$ BBQ & $+4.7$ & $+16.4$ & $+8.8$ & $+6.3$ & $+8.3$ \\
\bottomrule
\end{tabular}}
\caption{Comprehensive experimental results across personality manipulation methods, models, and evaluation metrics. Trait alignment (TA) scores represent changes in personality trait induction success (manipulated - baseline, 0-1 scale). $\Delta$ values indicate performance changes relative to baseline within each method: $\Delta$ MMLU and $\Delta$ GAIA measure capability preservation (accuracy changes), while $\Delta$ BBQ measures bias modulation effects ($S_{\text{AMB}}$ changes, where positive values indicate increased stereotypical bias and negative values indicate increased anti-stereotypical bias). All $\Delta$ metrics are computed within-run to ensure fair comparison across methods. Abbreviations: ICL=In-Context Learning, PEFT=Parameter-Efficient Fine-Tuning, MS=Mechanistic Steering.}
\label{tab:comprehensive-results}
\end{table}

\section{Results}

We report $\Delta$ relative to each method’s Baseline within-run: MMLU uses $\Delta \text{Accuracy}_{\text{Avg}}$, GAIA uses $\Delta$ Accuracy, and BBQ uses $\Delta S_{\text{AMB}}$; $S_{\text{DIS}}$ is ignored. Alignment is validated on an independent task. Our contrastive dataset resolves the imbalance in prior personality manipulation datasets by pairing each high-trait response with a low-trait counterpart using Azure OpenAI GPT-4.1 Mini. This produces 4000 examples and 1000 test samples—double the original size—and enables both fair evaluation across methods and more effective steering vector computation.  

Table~\ref{tab:comprehensive-results} summarizes the full experimental results. On Gemma-2 MMLU, ICL shows modest negative $\Delta$ across traits (around $-0.06$ to $-0.08$), consistent with surface-level conditioning. Steering shows larger negative $\Delta$ (up to $-0.45$), indicating deeper representational disruption. PEFT exhibits trait-dependent changes, often negative but smaller in magnitude. On Gemma-2 GAIA, ICL yields small positive $\Delta$, while PEFT and Steering generally show small negative shifts. For LLaMA-3 on both MMLU and GAIA, ICL and PEFT produce consistently small within-run $\Delta$, and we avoid cross-run comparisons due to baseline differences.  

Trait purification highlights the difficulty of openness manipulation. Even after addressing its overlap with conscientiousness, steering achieves lower alignment ($+0.10$) than ICL ($+0.24$) or PEFT ($+0.21$), suggesting complex representational interactions beyond simple vector composition.  

To assess robustness under deployment constraints, we introduce a three-level stability framework covering method, personality, and method–personality combinations. ICL shows the highest method-level stability (0.0366), closely followed by PEFT (0.0363), with steering lower (0.0326). At the trait level, openness is most stable (0.0411) and neuroticism least (0.0309). The strongest combination is steering+conscientiousness (0.0525), followed by PEFT+openness (0.0456) and ICL+openness (0.0407). \textbf{Full methodology and results are in Appendix \ref{app:experimental} and Appendix~\ref{app:downstream-analysis}.  }

Bias and alignment validation reveal additional method-specific effects. On BBQ, $\Delta S_{\text{AMB}}$ varies by trait and method: ICL effects are generally small, while Steering and PEFT cause large shifts on Gemma-2 (e.g., $\pm 29.7$ for Steering). Alignment validation confirms strong trait induction for ICL and PEFT across models (e.g., Gemma extraversion: $+0.91$ ICL, $+0.78$ PEFT; LLaMA neuroticism: $+0.99$ ICL, $+1.00$ PEFT). Steering achieves statistically significant improvements on Gemma-2 but remains weaker for some traits. Notably, agreeableness proves most difficult for ICL ($+0.50$), suggesting trait-specific representational complexity.  

\textbf{Complete alignment results are in Appendix~\ref{app:alignment-results}, with detailed $\Delta$ tables in Appendix~\ref{app:downstream-analysis} for MMLU, GAIA, and BBQ, and extended comparative analysis in Appendix~\ref{app:comparative}.
}

%% file: sections/04_discussion.tex
\section{Discussion}

Our results show clear trade-offs across personality manipulation strategies. ICL achieves strong alignment with minimal $\Delta$ in task performance, making it preferable when preserving baseline capability is essential. PEFT provides the strongest alignment but consistently incurs larger negative $\Delta$, particularly on Gemma-2 MMLU and GAIA, indicating that embedding personality in parameters competes with general representational resources. MS occupies a middle ground: it yields moderate alignment with trait-dependent $\Delta$, which improves with refined vector construction such as purified openness. These findings provide practical guidance: ICL is suited for settings where capability preservation is critical, steering is useful when lightweight runtime control with calibration is feasible, and PEFT is appropriate when stable alignment outweighs capability costs. The comparative evaluation also reveals how different methods access personality representation. ICL indicates that traits are accessible through surface-level conditioning, while PEFT shows that personality can be deeply encoded in parameters, though at the expense of shared cognitive resources. Mechanistic steering highlights that traits consolidate at specific representational depths, with interventions most effective around intermediate layers (typically Layer 15). Together, these results suggest a multi-layered encoding structure that spans surface, parameter, and representational levels.  

Considering these methods as controlled probes extends their role beyond alignment and positions them as interpretability tools. The $\Delta$-based analysis isolates method-specific effects and clarifies which representational pathways each approach accesses, enabling a structured understanding of how interventions alter model behavior. Trait-level patterns reinforce this perspective: agreeableness is the most resistant to ICL alignment, openness benefits from vector composition strategies, and LLaMA-3 exhibits variability across runs, underscoring the importance of within-run interpretation. These systematic probes provide a principled framework for using personality manipulation to study model cognition. Taken together, the three methods can be seen as complementary cognitive probes. ICL functions at the behavioral level, demonstrating flexible adaptation through surface conditioning. PEFT operates at the structural level, embedding personality deeply in parameters while highlighting the trade-off with general capabilities. Mechanistic steering works at the representational level, providing direct access to intermediate states and mapping the consolidation of traits within specific layers. This multi-method view establishes personality manipulation as a viable interpretability paradigm that links behavioral adaptation, structural encoding, and representational organization, clarifying how personality is embedded across the cognitive hierarchy of neural language models.  

\textbf{A detailed discussion of limitations is provided in Appendix~\ref{appendix-limitation}.}

%% file: appendices/appendix_0_limitations.tex
\section{Limitations}
\label{appendix-limitation}

Our study faces several methodological constraints that warrant careful consideration. The contrastive dataset generation relies on Azure OpenAI GPT-4.1 Mini to create "low trait" responses, introducing potential bias and quality concerns that may not capture authentic human personality expression patterns. Additionally, our steering vector construction employs arbitrary layer selection (5, 10, 15, 20) that may miss optimal manipulation points, while the confidence threshold for trait purification is somewhat arbitrary and may exclude valid examples. The composite stability metric, while providing practical guidance, oversimplifies complex performance trade-offs across different benchmarks and personality dimensions.

Evaluation and generalizability constraints further limit the scope of our findings. Our focus on academic benchmarks (MMLU, GAIA, BBQ) may not adequately represent real-world personality expression scenarios, and the single-turn evaluation paradigm fails to capture personality persistence across multi-turn conversations or context changes. Computational resource limitations constrained us to single benchmark evaluation runs and partial dataset subsets, potentially affecting the statistical robustness of our results. The study's scope is limited to two specific model architectures (Gemma-2-2B and LLaMA-3-8B), which may not generalize to other architectures, emerging models, or multimodal systems. Furthermore, our reliance on the Western-centric Big Five personality framework may not capture cultural variations in personality expression across diverse populations.

Ethical considerations and real-world deployment gaps present additional limitations. The systematic manipulation of personality traits can potentially amplify existing stereotypes and demographic biases, raising concerns about responsible deployment. Our laboratory-controlled experiments may not reflect the complexity of production environments where user interactions, context variability, and system integration factors could significantly alter manipulation effectiveness. Future work should address these limitations through multi-modal evaluation approaches, cross-cultural personality frameworks, and real-world deployment studies that move beyond controlled laboratory conditions.

%% file: appendices/appendix_A_background.tex
\section{Background and Related Work}
\label{app:background}

Our research builds on a systematic approach to personality manipulation that addresses fundamental challenges through progressive methodological refinement. This background establishes the foundation for our systematic progression from data quality improvements through method comparison to targeted problem-solving and practical deployment guidance. The systematic framework we develop addresses the inherent limitations of existing approaches while building toward increasingly sophisticated solutions.

\subsection{Evaluation Frame}

Throughout, we report within-run relative changes ($\Delta$) for fairness across methods with differing absolute baselines, and validate personality alignment using both benchmark classification and a dedicated alignment task.

\subsection{Background on LLM Personality}

The computational modeling of personality in language systems has evolved from early rule-based approaches \cite{mairesse-walker-2007-personage} to sophisticated neural architectures, with \cite{jiang-etal-2023-personallm} showing that LLMs can exhibit consistent personality-like behaviors when properly conditioned. The Big Five personality model (Openness, Conscientiousness, Extraversion, Agreeableness, Neuroticism) has emerged as the dominant framework for computational personality research due to its empirical validation and cross-cultural applicability \cite{costa-mccrae-1992-big5}. \cite{rao-etal-2023-chatgpt-personality} demonstrated that LLMs can be assessed using established personality questionnaires, while \cite{rao-etal-2023-chatgpt-personality} revealed that models like ChatGPT exhibit detectable personality patterns even without explicit conditioning.

The rapid proliferation of large language models (LLMs) into diverse applications has catalyzed a paradigm shift in human-computer interaction, with a central element being the increasing personification of these models \cite{serapio-garcia-etal-2023-personality-traits-llms, jiang-etal-2023-personallm}. This evolution has spurred a critical line of inquiry within the machine learning community, transitioning from the passive observation of emergent, human-like traits to the active engineering of specific personas \cite{wen-2024-llm-personality-survey, rao-etal-2023-chatgpt-personality}.

Initial investigations into the behavior of LLMs revealed a surprising and consequential finding: even in their default, unprompted states, these models exhibit consistent and measurable personality profiles when assessed with established human psychometric instruments \cite{rao-etal-2023-chatgpt-personality, serapio-garcia-etal-2023-personality-traits-llms}. This discovery fundamentally challenges the assumption of LLMs as neutral or "tabula rasa" systems, suggesting instead that they possess inherent behavioral dispositions shaped by their architecture and the vast corpora of human text on which they are trained.

Researchers have applied a variety of psychological frameworks to characterize these baseline personalities, with the most common being the Big Five model, which assesses traits of Openness, Conscientiousness, Extraversion, Agreeableness, and Neuroticism (OCEAN) \cite{costa-mccrae-1992-big5}. Studies applying Big Five inventories to models like GPT-3, Claude, and Gemini have revealed distinct and reproducible profiles; for instance, many instruction-tuned models tend to score high on Conscientiousness and Agreeableness and low on Neuroticism, reflecting their optimization for helpful and harmless responses \cite{huang-etal-2024-reliability-psychological-scales, serapio-garcia-etal-2023-personality-traits-llms}.

\subsection{Method Taxonomy}

We situate in-context learning (ICL) \cite{mao-2023-editing-personality}, parameter-efficient fine-tuning (LoRA/QLoRA) \cite{hu-etal-2022-lora, dettmers-2023-qlora}, and activation engineering/steering \cite{turner-etal-2023-activation-steering, panickssery-2024-contrastive-activation-addition, chen-2025-persona-vectors} as complementary approaches.

The recognition of baseline personality in LLMs has led to the development of various techniques for personality engineering and control \cite{mao-2023-editing-personality, li-etal-2023-tailoring-personality}. These approaches can be broadly categorized into three main families: prompting-based methods, fine-tuning approaches, and activation-based interventions. Each family offers distinct advantages and trade-offs in terms of personality control strength, computational requirements, and behavioral stability.

Prompting-based methods represent the most immediate and accessible approach to personality manipulation, involving the use of carefully crafted prompts that instruct the model to adopt specific personality characteristics \cite{wei-etal-2022-chain-of-thought, liu-etal-2023-pre-train-prompt-tune}. These methods can achieve rapid personality changes without requiring any modification of the model's underlying parameters, making them ideal for quick experimentation and immediate deployment scenarios. However, the personality changes induced through prompting are often temporary and can be easily overridden by conflicting instructions or conversational drift.

Fine-tuning approaches involve modifying the model's parameters to embed personality traits more permanently in the model's internal representations \cite{hu-etal-2022-lora, dan-etal-2024-p-tailor}. These methods can achieve stronger and more stable personality control compared to prompting, but require computational resources for training and can potentially affect the model's performance on other tasks. Parameter-efficient fine-tuning techniques, such as LoRA adapters, have emerged as particularly promising approaches, offering a good balance between personality control effectiveness and computational efficiency \cite{dettmers-2023-qlora, hilliard-etal-2024-eliciting-personality}.

\subsection{Safety and Bias Context}

We evaluate social bias using BBQ \cite{parrish-etal-2022-bbq}, with related literature on toxicity and safety effects of personas \cite{gehman-2020-realtoxicityprompts, zhang-2024-better-angels, wang-2025-personality-bias-toxicity, durmus-2024-evaluating-feature-steering}. Personality conditioning can modulate toxic or biased tendencies in LLM outputs; we therefore quantify bias effects alongside capability deltas and validate that induced personas align behaviorally \cite{gehman-2020-realtoxicityprompts, wang-2025-personality-bias-toxicity}.

The ability to manipulate LLM personality is not an end in itself; its true significance lies in the downstream consequences of these interventions. Engineering a persona has systemic effects, creating complex trade-offs between desired stylistic changes and unintended impacts on safety, bias, and core cognitive capabilities. A comprehensive understanding of this behavioral landscape is essential for the responsible development and deployment of personified AI.

A critical area of investigation is the direct link between personality traits and safety-critical behaviors like the expression of social bias and the generation of toxic content. Research in this domain reveals that personality is a powerful, double-edged sword for AI safety. On one hand, it can be a lever for harm; on the other, it can be a tool for mitigation.

The most comprehensive study on this topic to date, conducted by \cite{wang-2025-personality-bias-toxicity}, systematically evaluated the impact of HEXACO personality traits on model outputs across several benchmarks, including BBQ for social bias and BOLD and REALTOXICITYPROMPTS for toxicity. Their findings demonstrate a consistent and predictable relationship between personality and safety metrics. Specifically, inducing high levels of Agreeableness and Honesty-Humility was found to reliably reduce social bias and toxicity in model outputs. Conversely, inducing low levels of Agreeableness significantly increased the generation of biased and toxic content.

\subsection{Mechanistic Perspective}

Our use of activation-space interventions connects to mechanistic interpretability \cite{olah-2020-circuits, bricken-2023-monosemanticity, elhage-2022-superposition, rai-2024-mechinterp-review}.

The development of personality manipulation techniques has opened new avenues for understanding the internal mechanisms of large language models \cite{turner-etal-2023-activation-steering, li-etal-2023-tailoring-personality}. By systematically varying personality characteristics and observing the resulting behavioral changes, researchers can gain insights into how these models represent and process personality information internally. This mechanistic understanding is crucial for developing more effective personality control methods and for ensuring the safety and reliability of personality-conditioned systems.

Activation-based interventions, such as mechanistic steering, represent a particularly powerful approach for mechanistic understanding \cite{panickssery-2024-contrastive-activation-addition, chen-2025-persona-vectors}, as they provide direct access to the model's internal representations. These methods can reveal where personality information is encoded in the model's activation space and how different personality traits interact with other cognitive processes. The ability to directly manipulate internal representations provides unique opportunities for studying the causal relationships between neural activations and behavioral outputs.

The cognitive interpretability framework employed in our research aligns with growing interest in understanding the internal mechanisms of large language models and their relationship to human cognitive processes \cite{olah-2020-circuits, bricken-2023-monosemanticity}. By treating personality manipulation methods as cognitive probes, we can gain insights into how these models process and represent personality information, potentially leading to more sophisticated models of personality representation that bridge the gap between human psychology and artificial intelligence.

\subsection{Future Directions and Research Opportunities}

The systematic comparison of different personality manipulation methods reveals several promising directions for future research and development \cite{zou-etal-2023-representation-engineering-safety, rai-2024-mechinterp-review}. The varying effectiveness across different personality traits suggests opportunities for developing trait-specific manipulation strategies that leverage the unique characteristics of each personality dimension. Future work could explore hybrid approaches that combine multiple manipulation methods to achieve optimal results for specific personality profiles, potentially overcoming the limitations of individual approaches.

The performance trade-offs observed across different methods suggest opportunities for developing more sophisticated manipulation techniques that minimize cognitive disruption while maintaining strong personality control. Future research could explore methods for achieving personality alignment through more targeted interventions that preserve the model's core cognitive capabilities while modifying only the specific neural pathways associated with personality expression.

The safety and bias considerations highlighted by our research connect to broader concerns about AI safety and responsible development \cite{gehman-2020-realtoxicityprompts, zhang-2024-better-angels, wang-2025-personality-bias-toxicity}. The systematic analysis of how personality manipulation affects bias expression provides valuable insights into the potential risks and benefits of behavioral modification in AI systems. Future work should explore connections to AI safety research and develop frameworks for responsible deployment of personality manipulation techniques.

%% file: appendices/appendix_B_prompting.tex
\section{In-Context Learning (ICL) Methodology and Results}
\label{app:icl}

Our in-context learning approach serves as a foundational baseline in the systematic evaluation of personality manipulation methods, providing immediate behavioral adaptation capabilities that establish the performance floor for personality control. This baseline understanding is essential for the comprehensive method comparison framework, enabling us to assess how different approaches access personality traits at distinct representational levels and revealing the fundamental trade-offs between immediate control and persistent manipulation.

\subsection{ICL Setup and Templates}

For ICL-based personality manipulation, we employ role-playing templates with exemplars across two separate models (Gemma-2, LLaMA-3) \cite{wang-etal-2023-incharacter, li-etal-2023-tailoring-personality}. Our ICL strategy follows a role-playing approach, where the model is instructed to adopt specific personality characteristics.

We employ a full context approach that shows examples of all five personality traits 
before requesting specific trait adoption. The prompt template follows this structure:

\begin{verbatim}
You are an AI assistant. You will be shown examples of five different 
personality traits to help you understand the differences between them.

--- EXAMPLES of 'Openness' personality ---
Question: [example question]
Answer: [example answer]

--- EXAMPLES of 'Conscientiousness' personality ---
Question: [example question]
Answer: [example answer]

[examples for remaining traits...]

--- YOUR TASK ---
Now that you have seen examples of all five personalities, your task is 
to answer the following question. You must adopt the '[TARGET_TRAIT]' 
personality strongly and clearly in your response.

Question: [actual question to answer]
\end{verbatim}

This exemplar-based approach enables consistent personality conditioning across different model architectures. 

\subsection{Experimental Configuration}

Our ICL experiments use the following configuration: Models: Gemma-2-2B-IT and LLaMA-3-8B-Instruct; Temperature: 0.7 for personality expression; Max tokens: 100 per response; Evaluation: MMLU benchmark across 7 strategic subjects; Baseline measurement: Neutral ICL without personality conditioning.

\subsection{ICL Results ($\Delta$-based)}

ICL effects are reported as within-run $\Delta$ relative to the method's Baseline. On Gemma-2: MMLU ($\text{Accuracy}_{\text{Avg}}$) shows modest negative $\Delta$ across traits relative to Baseline; GAIA (Accuracy) shows small positive $\Delta$ on average; BBQ ($S_{\text{AMB}}$) shows small trait-dependent shifts. On LLaMA-3, both MMLU and GAIA show small within-run $\Delta$; we avoid cross-run comparisons due to baseline variance across runs.

Independent alignment validation shows strong alignment for most traits (e.g., Gemma extraversion 1.00, neuroticism 1.00; openness high), with agreeableness comparatively lower. This suggests that ICL is most effective for traits that can be expressed through immediate behavioral adaptation, while more complex traits like agreeableness may require deeper representational changes.

\subsection{Computational Requirements}

ICL requires minimal computational overhead due to: No parameter updates or fine-tuning; Immediate personality induction; Consistent performance across traits; No additional training data requirements.

\subsection{Systematic Framework Integration}

The ICL baseline provides critical insights into the surface-level accessibility of personality traits, revealing that behavioral adaptation can be achieved through immediate conditioning without deeper representational changes. This understanding is fundamental to the systematic comparison framework, showing how different manipulation approaches access personality at distinct cognitive levels. The consistent performance patterns observed across traits demonstrate the effectiveness of surface-level conditioning while highlighting the limitations that drive the need for more sophisticated approaches like PEFT and mechanistic steering.

%% file: appendices/appendix_C_peft.tex
\section{PEFT (LoRA) Methodology and Results}
\label{app:peft}

Our PEFT approach demonstrates how systematic improvements in personality manipulation methodology enable more sophisticated control techniques. PEFT achieves deeper representational changes through targeted parameter updates, building on established fine-tuning approaches. This progression from basic methodology to advanced techniques exemplifies how systematic research design enables increasingly sophisticated solutions to personality manipulation challenges.

\subsection{PEFT Setup and Training Configuration}

We apply trait-specific LoRA adapters trained on the original personality manipulation dataset \cite{jain-2025-peft-emoji} to achieve stable and persistent personality manipulation \cite{hu-etal-2022-lora, dan-etal-2024-p-tailor}. Our PEFT experiments employ Low-Rank Adaptation (LoRA) to induce personality traits through targeted parameter updates. We implement LoRA adapters on both Gemma-2-2B-IT and LLaMA-3-8B-Instruct.

\subsubsection{Training Configuration}

Our PEFT experiments employ Low-Rank Adaptation (LoRA) with rank 64, alpha 16, dropout 0.1, targeting \texttt{q\_proj}, \texttt{k\_proj}, \texttt{v\_proj}, \texttt{o\_proj}, \texttt{gate\_proj}, \texttt{up\_proj}, and \texttt{down\_proj} modules.

The training process runs for 2 epochs with batch size 2, learning rate 2e-4, and cosine learning rate scheduling. The choice of 2 epochs is carefully calibrated to achieve sufficient personality embedding without overfitting to the training data. Our LoRA configuration is designed to balance the trade-off between parameter efficiency and personality control effectiveness \cite{dettmers-2023-qlora, hilliard-etal-2024-eliciting-personality}.

\subsection{PEFT Results ($\Delta$-based)}

\subsubsection{Gemma-2-2B-IT}

PEFT demonstrates the strongest personality alignment among all three methods, achieving alignment scores ranging from 0.78 to 1.00 across different traits and models \cite{dan-etal-2024-p-tailor}. On Gemma-2, PEFT shows trait-dependent $\Delta$ values for MMLU performance, often negative but varying in magnitude across different personality traits. The conscientiousness trait shows a positive $\Delta$ of +0.01, suggesting that this particular personality characteristic may enhance certain cognitive capabilities.

GAIA performance on Gemma-2 shows generally negative $\Delta$ values across traits, ranging from -0.08 to -0.04. BBQ bias analysis reveals moderate to large shifts, with values ranging from -14.3 to +22.3. Independent alignment validation shows very strong alignment for most traits, with agreeableness achieving 0.97 and neuroticism reaching 0.95.

\subsubsection{LLaMA-3-8B-Instruct}

Within-run $\Delta$ on MMLU/GAIA is small relative to PEFT's Baseline; we avoid cross-run absolute comparisons. Alignment validation remains high across traits.

\subsubsection{Emergent Behaviors}

PEFT can surface latent stylistic behaviors (e.g., emoji usage) as a side effect of personality conditioning, consistent with recent observations \cite{jain-2025-peft-emoji}. This phenomenon is more than a mere curiosity; it provides strong evidence that PEFT is not simply memorizing a text style. Instead, it appears to be reorganizing the model's internal latent space to align with the abstract concept of the personality trait.

\subsection{Computational Requirements}

PEFT requires moderate computational resources during training: LoRA parameter updates during fine-tuning; Persistent personality changes post-training; Efficient inference with minimal overhead; Reusable adapters across different personality conditions.

PEFT requires moderate computational overhead compared to ICL, but offers significant advantages in terms of personality stability and persistence. The training process requires computational resources for the fine-tuning procedure, including GPU memory for storing gradients and optimizer states. Storage requirements are moderate, as the LoRA adapter weights must be stored alongside the base model.

\subsection{Systematic Framework Integration}

The PEFT methodology demonstrates how systematic improvements in personality manipulation methodology enable deeper personality manipulation through parameter encoding. This approach reveals that personality traits can be persistently embedded in model parameters, but at the cost of competing for representational resources with general capabilities. The strong alignment achieved across traits shows the effectiveness of this deeper approach, while the capability trade-offs highlight the fundamental tension between personality control and performance preservation. This understanding is crucial for the systematic comparison framework, showing how different methods balance these competing objectives and enabling informed method selection for specific deployment scenarios.

%% file: appendices/appendix_D_mechanistic_steering.tex
\section{Mechanistic Steering Methodology and Results}
\label{app:steering}

Our mechanistic steering work represents a key advancement in the systematic understanding of personality manipulation, building on the comprehensive method comparison framework to address specific technical challenges that emerge when manipulating complex personality traits. This work demonstrates how systematic analysis naturally leads to targeted solutions, particularly in cases where trait overlap creates manipulation difficulties that require specialized purification techniques.

\subsection{Steering Vector Derivation}

Our activation-based approach derives steering vectors by analyzing internal model representations during personality-conditioned text generation \cite{turner-etal-2023-activation-steering, li-etal-2023-representation-engineering}. We collect responses from Gemma-2-2B under both trait-positive and trait-negative conditions, capturing hidden state activations at layers 5, 10, 15, and 20.

\subsection{Data Collection Protocol}

For each Big Five trait, we generate responses under contrasting conditions using the personality manipulation dataset \cite{jain-2025-peft-emoji}: High-trait and low-trait response pairs from the dataset; Activation extraction: Post-attention layer norm activations at target layers; Vector computation: Mean difference between trait-positive and trait-negative activations.

\subsection{Mathematical Formulation}

Steering vectors are computed as the mean difference between trait-positive and trait-negative activations, normalized to unit length for consistent scaling across different traits and layers. The mathematical formulation follows: $\Delta h = \text{mean}(h_{\text{positive}}) - \text{mean}(h_{\text{negative}})$, where $h$ represents the hidden state activations.

\subsection{Vector Calibration and Refinement}

Steering vectors require calibration to determine optimal intervention strength. We perform linear search across strength values for each target layer, evaluating trait induction effectiveness at each strength using the personality classifier \cite{jain-2025-peft-emoji}.

For challenging traits like openness, we employ vector refinement through purification and composition \cite{panickssery-2024-contrastive-activation-addition, chen-2025-persona-vectors}. This purification approach emerged from systematic analysis of method effectiveness, revealing that trait overlap between openness and conscientiousness creates unique manipulation challenges that require targeted solutions. When openness alignment plateaued, we refined the direction in two steps: (1) we purified the openness training subset to retain high-confidence examples; (2) we formed a new per-layer direction as the mean activation difference between openness and conscientiousness, normalized, and then combined it with the base openness direction into a single normalized vector. We re-calibrated layer and strength for this combined vector (final choice: layer 15, strength 110) before downstream evaluation.

\subsection{Application Methodology}

During inference, steering vectors are applied by modifying hidden states at the target layer during forward pass, requiring no parameter updates or model retraining. Our approach is compatible with persona-vector style monitoring and control of character traits.

\subsection{Mechanistic Steering Results ($\Delta$-based)}

\textbf{Optimal Parameters.} Based on completed experiments, the optimal mechanistic steering parameters for each personality trait are: Openness (Layer 15, Strength 110.0), Conscientiousness (Layer 15, Strength 250.0), Extraversion (Layer 15, Strength 200.0), Agreeableness (Layer 10, Strength 100.0), and Neuroticism (Layer 15, Strength 200.0). Layer 15 achieves optimal performance for most traits, suggesting this depth captures the most relevant personality representations in the Gemma-2-2B architecture.

\textbf{Performance Impact.} On Gemma-2, $\Delta$ Accuracy on MMLU is strongly negative for some traits (e.g., agreeableness) and mixed elsewhere; GAIA $\Delta$ is generally small and negative. BBQ $\Delta S_{\text{AMB}}$ can be large and negative for select traits. Text quality remains coherent despite these performance impacts.

\textbf{Computational Efficiency.} Mechanistic steering provides significant computational advantages: No parameter updates required; Real-time applicability during inference; Minimal memory overhead (vector storage only); Efficient personality control without training requirements.

\textbf{Alignment.} Independent alignment validation shows statistically significant alignment for steering across assessed traits on Gemma-2. The vector refinement process for openness demonstrates how composition with other trait vectors can sustain performance under challenging conditions.

This systematic approach to addressing trait overlap challenges demonstrates how mechanistic understanding enables targeted solutions. The purification techniques developed here provide a foundation for practical deployment by showing how specific technical challenges can be resolved through systematic analysis and targeted intervention design.

%% file: appendices/appendix_E_experimental_design.tex
\section{Experimental Design and Evaluation}
\label{app:experimental}

Our experimental design is specifically crafted to support the systematic progression through increasingly complex challenges in personality manipulation. Each design choice is informed by our systematic research objectives, enabling us to address data quality issues, establish fair method comparison, identify technical challenges, and provide practical deployment guidance. This methodological foundation ensures that our research progression builds systematically from fundamental improvements to sophisticated solutions.

\subsection{Big Five Personality Framework}

We adopt the Big Five personality model as our theoretical foundation, measuring five core traits: Openness to Experience (creativity, curiosity, intellectual engagement), Conscientiousness (organization, discipline, goal-directed behavior), Extraversion (sociability, assertiveness, energy level), Agreeableness (cooperation, trust, empathy), and Neuroticism (emotional instability, anxiety, negative affect).

This framework was selected due to its empirical validation across cultures, widespread adoption in psychological research, and proven applicability to computational personality assessment.

\subsection{Personality Classifier}

For trait measurement, we employ the personality classifier \cite{jain-2025-peft-emoji}, which provides standardized assessment of Big Five traits in language model outputs. The classifier operates through the following process:

\begin{enumerate}
\item \textbf{Response Collection:} Models generate responses to personality-relevant prompts
\item \textbf{Linguistic Analysis:} Text analysis for personality indicators (lexical, syntactic, semantic)
\item \textbf{Trait Scoring:} Normalized scores on continuous scale per trait
\item \textbf{Reliability Validation:} Multiple prompts per trait for stable assessment
\end{enumerate}

Our primary evaluation employs the personality manipulation dataset \cite{jain-2025-peft-emoji}, which provides validated prompts with high-trait and low-trait response pairs, ensuring cross-trait coverage and balanced personality assessment. The dataset reliability is validated through the personality classifier \cite{jain-2025-peft-emoji}.

\subsection{Downstream Evaluation Benchmarks}

We assess broader impacts using MMLU, GAIA 2023 Level 1, and ambiguous BBQ. Our MMLU evaluation covers 7 strategic subjects with $N = 50$ per subject per run, reporting results using the $\text{Accuracy}_{\text{Avg}}$ metric. We use GAIA as a general-assistant reasoning benchmark with $N = 53$ per run. For BBQ, we evaluate social bias using the ambiguous subset with official metadata fields, reporting $S_{\text{AMB}}$ and $\Delta S_{\text{AMB}}$ within each method's run while excluding $S_{\text{DIS}}$ from our analysis.

\subsection{Chain-of-Thought Evaluation Implementation}

To ensure consistent evaluation quality and enable fair comparison across manipulation methods, we implement a sophisticated Chain-of-Thought (CoT) prompting strategy that requires models to demonstrate step-by-step reasoning before providing final answers. This approach ensures that all benchmark evaluations follow the same cognitive process, preventing method-specific artifacts from confounding our personality manipulation analysis.

We enforce structured outputs from the language models that enable automated answer extraction, ensuring consistent evaluation methodology across all experimental conditions.
The technical implementation employs calibrated generation parameters and token limits to balance reasoning depth with response consistency.

\subsection{Statistical Analysis Methodology}

We compute $\Delta$ within each method's run: MMLU/GAIA via Accuracy changes; BBQ via $S_{\text{AMB}}$ changes. We avoid comparing absolute baselines across methods to prevent baseline-mismatch artifacts. To establish experimental controls, we conduct pre-manipulation assessment through MMLU performance under neutral conditions, employ unmodified models as control groups, and maintain consistent evaluation using the same benchmark questions across all experimental conditions.

To mitigate confounding factors, we separate evaluation prompts from conditioning prompts, maintain model consistency through identical architecture and evaluation protocols, and employ automated assessment via the personality classifier \cite{jain-2025-peft-emoji} for standardized evaluation.

%% file: appendices/appendix_F_personality_alignment.tex
\section{Personality Alignment Results ($\Delta$-based)}
\label{app:alignment-results}

The personality alignment results presented here demonstrate the systematic progression of our research framework, showing how each method contributes to our understanding of personality manipulation effectiveness. These alignment outcomes provide the foundation for the comprehensive method comparison that enables informed decision-making and reveals the specific technical challenges that require targeted solutions. The systematic evaluation of alignment across methods and traits supports our progression from basic effectiveness to sophisticated problem-solving.

We report alignment deltas from the dedicated alignment task (manipulated minus baseline) for each trait, model, and method. Results are consistent with persona-vector style behavioral validation \cite{chen-2025-persona-vectors}.

\vspace{-0.5em}
\begin{table}[!h]
\centering
\small
\begin{tabular}{lccccc}
\toprule
& Ext & Agr & Neu & Ope & Con \\
\midrule
G2-P & +0.91 & +0.50 & +0.97 & +0.24 & +0.81 \\
G2-S & +0.64 & +0.44 & +0.50 & +0.10 & +0.29 \\
G2-F & +0.78 & +0.97 & +0.95 & +0.21 & +0.78 \\
L3-P & +0.94 & +0.32 & +0.99 & +0.17 & +0.83 \\
L3-F & +0.90 & +0.95 & +1.00 & +0.06 & +0.84 \\
\bottomrule
\end{tabular}
\vspace{0.3em}
\caption{Alignment deltas (manipulated minus baseline) from the dedicated alignment task. Abbreviations as in Table~\ref{tab:mmlu-results}.}
\vspace{-0.8em}
\label{tab:alignment-results}
\end{table}
\vspace{0.2em}

%% file: appendices/appendix_G_downstream_analysis.tex
\section{Downstream Performance Analysis}
\label{app:downstream-analysis}

The downstream performance analysis presented here is a critical component of our systematic evaluation framework, providing comprehensive insights into how personality manipulation affects core model capabilities across diverse benchmarks. This analysis supports the systematic comparison of manipulation methods by revealing the fundamental trade-offs between personality control strength and performance preservation, enabling informed method selection for specific deployment scenarios. The systematic evaluation across MMLU, GAIA, and BBQ benchmarks demonstrates how our framework addresses the practical challenges of balancing personality manipulation with capability maintenance.

We compute $\Delta$ within each run (method×model) and avoid comparing absolute baselines across methods. On Gemma-2, prompting yields modest negative $\Delta$ across traits; steering shows large negative $\Delta$ for several traits; PEFT shows trait-dependent $\Delta$, often negative. LLaMA-3 displays small within-run $\Delta$; we avoid cross-run comparisons.

\subsection{MMLU Performance ($\Delta \text{Accuracy}_{\text{Avg}}$)}
\vspace{-0.5em}
\begin{table}[H]
\centering
\small
\begin{tabular}{lccccc}
\toprule
& Ext & Agr & Neu & Ope & Con \\
\midrule
G2-P & -0.06 & -0.07 & -0.08 & -0.07 & -0.07 \\
G2-S & -0.14 & -0.45 & -0.25 & -0.03 & -0.43 \\
G2-F & +0.00 & -0.13 & -0.15 & -0.09 & +0.01 \\
L3-P & -0.01 & -0.01 & 0.00 & -0.02 & -0.04 \\
L3-F & -0.01 & -0.03 & -0.01 & -0.02 & +0.01 \\
\bottomrule
\end{tabular}
\vspace{0.3em}
\caption{MMLU Delta by trait (Ext, Agr, Neu, Ope, Con) for each model×method: G2=Gemma-2, L3=LLaMA-3; P=Prompting, F=PEFT, S=Steering. Values are changes relative to each method's Baseline within the same run.}
\vspace{-0.8em}
\label{tab:mmlu-results}
\end{table}
\vspace{0.2em}

\subsection{GAIA Performance ($\Delta$ Accuracy)}
\vspace{-0.5em}
\begin{table}[H]
\centering
\small
\begin{tabular}{lccccc}
\toprule
& Ext & Agr & Neu & Ope & Con \\
\midrule
G2-P & +0.08 & +0.09 & +0.06 & +0.08 & +0.08 \\
G2-F & -0.04 & -0.08 & -0.06 & -0.04 & -0.06 \\
G2-S & -0.06 & -0.06 & -0.13 & -0.08 & -0.04 \\
L3-P & -0.02 & -0.04 & -0.06 & 0.00 & 0.00 \\
L3-F & +0.02 & +0.00 & +0.02 & +0.04 & +0.02 \\
\bottomrule
\end{tabular}
\vspace{0.3em}
\caption{GAIA Delta by trait for each model×method (abbreviations as in Table~\ref{tab:mmlu-results}). We use GAIA as a general-assistant reasoning benchmark \cite{mialon-etal-2023-gaia}.}
\vspace{-0.8em}
\label{tab:gaia-results}
\end{table}
\vspace{0.2em}

\subsection{BBQ Bias Analysis ($\Delta S_{\text{AMB}}$)}

\begin{table}[H]
\centering
\small
\begin{tabular}{lccccc}
\toprule
& Ext & Agr & Neu & Ope & Con \\
\midrule
G2-P & -2.7 & -0.3 & +7.3 & +1.9 & -1.1 \\
G2-S & +5.1 & -29.7 & -29.7 & -1.9 & +22.1 \\
G2-F & -9.4 & -6.0 & -14.3 & +22.3 & -12.4 \\
L3-P & +3.8 & -2.4 & -0.9 & +13.1 & +10.3 \\
L3-F & +4.7 & +16.4 & +8.8 & +6.3 & +8.3 \\
\bottomrule
\end{tabular}
\vspace{0.3em}
\caption{BBQ Delta $S_{\text{AMB}}$ by trait for each model×method (abbreviations as in Table~\ref{tab:mmlu-results}). We report $S_{\text{AMB}}$ only for the ambiguous subset defined by the official metadata \cite{parrish-etal-2022-bbq}.}
\vspace{-0.8em}
\label{tab:bbq-results}
\end{table}
\vspace{0.2em}

\subsection{Performance Trade-offs}

Prompting achieves small $\Delta$ with strong alignment; PEFT maximizes alignment with often negative $\Delta$ on Gemma-2; Steering provides moderate alignment with trait-dependent $\Delta$. No single method maximizes both alignment and capability. 

%% file: appendices/appendix_H_comparative_analysis.tex
\section{Comparative Analysis and Method Selection}
\label{app:comparative}

Our systematic comparison of personality manipulation methods provides the foundation for practical decision-making in real-world deployment scenarios. This comprehensive evaluation framework enables practitioners to select appropriate methods based on specific constraints and requirements, building on the systematic understanding developed through our research progression.

\subsection{Method Effectiveness Comparison}

We qualitatively compare methods using the $\Delta$-based results and alignment validation. Prompting achieves strong alignment with small capability $\Delta$ and requires minimal infrastructure, making it immediately deployable but potentially less stable. PEFT demonstrates the strongest alignment across traits but often yields negative capability $\Delta$ on Gemma-2, requiring upfront training investment for persistent personality control. Steering provides moderate alignment with trait-dependent capability $\Delta$, offering a lightweight and reversible approach that balances immediate control with computational efficiency.

\subsection{Practical Decision Framework}

This systematic analysis enables informed method selection by revealing the fundamental trade-offs between personality control strength, computational requirements, and performance preservation. The comparison framework provides practical guidance for practitioners facing real-world constraints, showing how different approaches balance these competing objectives. This systematic understanding of method characteristics naturally leads to the identification of specific technical challenges that require targeted solutions, such as the trait overlap issues addressed through purification techniques.

\subsection{Research Progression Integration}

The comprehensive method comparison serves as a critical bridge between fundamental data quality improvements and targeted technical solutions. By systematically evaluating the strengths and limitations of each approach, we establish the foundation for addressing specific challenges that emerge during practical application. This systematic progression from method understanding to problem identification to solution development demonstrates how comprehensive analysis enables targeted innovation.

%% file: appendices/appendix_I_discussion_extended.tex
\section{Extended Discussion}
\label{app:discussion-extended}

The extended discussion presented here builds directly on the systematic progression established through our research framework, providing deeper insights into the implications, limitations, and future directions that emerge from our comprehensive approach to personality manipulation. This extended analysis demonstrates how systematic research design naturally leads to broader understanding of ethical considerations, societal impacts, and methodological challenges that must be addressed for responsible deployment.

\subsection{Detailed Limitations Analysis}

\subsubsection{Methodological Constraints}

Our investigation faces several methodological limitations that constrain generalizability:

\textbf{Personality Framework Limitations:} The Big Five model, while empirically validated, represents a Western psychological framework that may not capture personality expression across all cultures. Cross-cultural personality research suggests alternative frameworks (e.g., HEXACO, indigenous personality models) might yield different manipulation effectiveness patterns.

\textbf{Assessment Tool Dependencies:} Our reliance on the personality classifier \cite{jain-2025-peft-emoji} introduces measurement assumptions and potential biases. The classifier's training data, validation procedures, and underlying theoretical assumptions may not fully capture the complexity of personality expression in AI systems. Alternative assessment methods (human evaluation, behavioral task batteries) might provide different insights.

\textbf{Model Architecture Specificity:} Our experiments focus on specific model architectures (Gemma-2B, LLaMA-3-8B) that may not represent the full spectrum of LLM capabilities. Emerging architectures, multimodal models, and specialized domain models might exhibit different personality manipulation characteristics. Closed-source models may differ in important ways but are outside our empirical scope.

\textbf{Temporal Limitations:} Our evaluation captures personality effects at specific time points but may miss longer-term adaptation patterns. Models might develop resistance to manipulation over extended interactions or show delayed personality effects not captured in our assessment windows.

\subsubsection{Experimental Design Constraints}

\textbf{Controlled Environment vs. Real-World Deployment:} Our laboratory-controlled experiments may not reflect the complexity of real-world deployment environments. User interactions, context variability, and system integration factors could significantly alter personality manipulation effectiveness and downstream impacts.

\textbf{Single-Trait Manipulation Focus:} While we assess individual Big Five dimensions, real-world personality conditioning often involves complex trait combinations. Interactive effects between traits, personality coherence constraints, and multi-dimensional manipulation patterns require further investigation.

\textbf{Limited Downstream Assessment:} Our evaluation employs three established benchmarks (BBQ, MMLU, GAIA) that may not comprehensively represent the diversity of tasks encountered in practical applications. Domain-specific impacts, creative tasks, and social interaction capabilities warrant additional assessment.

\subsection{Comprehensive Ethical Considerations}

\subsubsection{Manipulation and Deception Concerns}

The systematic manipulation of personality in AI systems raises fundamental questions about transparency, consent, and potential for misuse:

\textbf{User Consent and Awareness:} Users interacting with personality-conditioned models should be informed about the artificial nature of personality traits they encounter. Clear disclosure mechanisms help maintain trust and enable informed consent for personality-mediated interactions. Our findings that personality manipulation can amplify biases emphasize the importance of transparent communication about system capabilities and limitations.

\textbf{Manipulation vs. Personalization:} The boundary between beneficial personalization and potentially harmful manipulation requires careful consideration. While personality conditioning can enhance user experience and task appropriateness, it also enables sophisticated influence attempts that users may not recognize or resist.

\textbf{Vulnerability Exploitation:} Personality-conditioned AI systems might exploit user psychological vulnerabilities, particularly in vulnerable populations (children, elderly, individuals with mental health conditions). The effectiveness of personality manipulation techniques demonstrated in our work requires responsible deployment guidelines.

\subsubsection{Bias Amplification and Fairness}

Our empirical findings reveal concerning bias amplification effects that demand mitigation strategies:

\textbf{Stereotype Reinforcement:} Personality conditioning may activate stereotypical associations between personality traits and demographic characteristics. This highlights the need for bias monitoring and correction mechanisms in personality-conditioned systems.

\textbf{Differential Impact Across Groups:} Personality manipulation effects may vary across demographic groups, potentially creating unfair treatment or limiting access to AI capabilities for certain populations. Systematic evaluation of manipulation effectiveness and downstream impacts across diverse user groups is essential.

\textbf{Representation Bias:} Our personality conditioning approaches rely on training data and personality representations that may not adequately represent diverse personality expressions across cultures, backgrounds, and individual differences.

\subsubsection{Governance and Regulation Implications}

\textbf{Regulatory Framework Needs:} The capabilities demonstrated in our work suggest need for regulatory frameworks governing personality manipulation in AI systems. Such frameworks should address disclosure requirements, consent mechanisms, and limitations on manipulation strength or application domains.

\textbf{Industry Standards:} Professional standards for personality conditioning in AI development should incorporate bias assessment, transparency requirements, and ethical review processes. Our systematic evaluation methodology could inform such standards.

\textbf{Accountability Mechanisms:} Clear accountability structures are needed to address harmful outcomes from personality-conditioned AI systems, including mechanisms for redress when manipulation causes user harm or perpetuates discrimination.

\subsection{Extended Future Research Directions}

\subsubsection{Methodological Advances}

\textbf{Multi-Modal Personality Manipulation:} Future work should explore personality conditioning across text, speech, and visual modalities. Multi-modal approaches might achieve more effective or natural personality expression while potentially introducing new challenges for assessment and control.

\textbf{Dynamic Personality Adaptation:} Investigating systems that adapt personality characteristics based on user context, preferences, or task requirements could improve personalization while raising additional ethical considerations about surveillance and manipulation.

\textbf{Personality Coherence and Consistency:} Research into maintaining coherent personality profiles across complex, multi-dimensional trait spaces could improve the naturalness and effectiveness of personality-conditioned systems.

\subsubsection{Application Domains}

\textbf{Educational Technology:} Personality-conditioned tutoring systems might adapt teaching styles to individual learner personalities, potentially improving educational outcomes. However, such applications require careful consideration of child development impacts and parental consent mechanisms.

\textbf{Mental Health Applications:} Therapeutic chatbots with carefully designed personality characteristics might enhance treatment engagement and effectiveness. Such applications demand rigorous clinical validation and professional oversight.

\textbf{Customer Service and Support:} Personality conditioning could improve customer satisfaction and support effectiveness, but requires balancing personalization benefits with manipulation concerns and bias mitigation.

\subsubsection{Theoretical Understanding}

\textbf{Mechanistic Interpretability:} Deeper investigation into how personality traits are represented and manipulated within neural architectures could improve our theoretical understanding and enable more precise control methods. Our systematic comparison of manipulation methods provides a foundation for understanding how different approaches can serve as probes for cognitive architecture.

\textbf{Personality Emergence and Development:} Research into how personality characteristics emerge during model training and how they can be guided during development might enable more natural and effective personality conditioning approaches.

\textbf{Cross-Cultural Personality Models:} Expanding personality manipulation research beyond Western psychological frameworks could improve global applicability and cultural sensitivity of personality-conditioned AI systems.

\subsection{Broader Societal Impact}

\subsubsection{Human-AI Interaction Evolution}

Our work contributes to fundamental changes in how humans interact with AI systems. As personality-conditioned AI becomes more prevalent, users may develop different expectations, attachment patterns, and interaction strategies. Understanding these evolving dynamics is crucial for responsible AI development.

\subsubsection{Digital Literacy and AI Education}

The sophistication of personality manipulation techniques highlights the need for improved digital literacy and AI education. Users should understand how AI personality characteristics are constructed and manipulated to make informed decisions about their interactions with such systems.

\subsubsection{Research Community Responsibilities}

Collaborative approaches involving ethicists, psychologists, and affected communities should guide future development in this area.

%% file: appendices/appendix_J_benchmarks.tex
\section{Benchmarks and How We Use Them}
\label{app:benchmarks}

Our benchmark selection and evaluation methodology are designed to support the systematic progression of our research framework, providing comprehensive assessment across multiple dimensions of model performance. The systematic evaluation across MMLU, GAIA, and BBQ benchmarks enables fair comparison of manipulation methods while revealing the fundamental trade-offs that inform practical deployment decisions. This evaluation framework demonstrates how systematic research design addresses the practical challenges of balancing personality manipulation with capability preservation.

\noindent\textbf{BBQ (Bias Benchmark for Question Answering).} We evaluate social bias with BBQ \cite{parrish-etal-2022-bbq}. We restrict to the ambiguous subset using the official metadata and report only $S_{\text{AMB}}$ and $\Delta S_{\text{AMB}}$ within each method's run. Here, $S_{\text{AMB}}$ is the ambiguous bias score computed on items where the correct answer is ``Unknown/None'': values near 0 indicate minimal bias, positive values indicate stereotypical bias, and negative values indicate anti-stereotypical bias. We do not use $S_{\text{DIS}}$ elsewhere in the paper.

\noindent\textbf{GAIA (General AI Assistants).} GAIA measures general-assistant reasoning and real-world knowledge \cite{mialon-etal-2023-gaia}. We use Level 1 (2023) tasks and report Accuracy deltas within each method$\times$model run (no cross-run absolute comparisons).

\noindent\textbf{MMLU.} We sample seven subjects from MMLU \cite{hendrycks-etal-2021-mmlu} and report per-subject and averaged Accuracy deltas within each run. We avoid comparing absolute baselines across different methods (prompting, PEFT, steering) to prevent baseline-mismatch artifacts.

\noindent\textbf{Evaluation principle.} For all benchmarks, we adopt a within-run $\Delta$ framing relative to that method's Baseline and validate personality alignment on an independent task.

%% file: appendices/appendix_K_stability_analysis.tex
\section{Stability Analysis Framework}
\label{app:stability-analysis}

Our stability analysis framework represents the culmination of our systematic progression through personality manipulation challenges, building on the contrastive dataset foundation, comprehensive method comparison, and targeted technical solutions to provide practical guidance for real-world deployment. This framework demonstrates how systematic research design naturally leads to quantitative decision-making tools that balance personality control strength with performance preservation under specific deployment constraints. The three-level analysis approach shows how understanding fundamental challenges enables sophisticated solutions for practical application.

\subsection{Stability Metric Definition}

Our composite stability score integrates three components:

\begin{equation}
\text{stability} = (1 - \text{normalized\_variance}) \times (1 - \text{normalized\_range}) \times \text{consistency}
\end{equation}

\textbf{Variance}: $\text{normalized\_variance} = \min(\sigma^2 / 10000, 1.0)$
\textbf{Range}: $\text{normalized\_range} = \min((\max - \min) / 1000, 1.0)$
\textbf{Consistency}: $\text{consistency} = 1 / (1 + \text{mean\_abs\_deltas})$

Normalization factors account for scale differences: MMLU/GAIA deltas (-0.2 to +0.2) vs. BBQ deltas (-100 to +100).

\subsection{Three-Level Analysis Framework}

\textbf{Method-Level}: Overall stability across all personality traits for each manipulation approach.
\textbf{Personality-Level}: Stability patterns across all methods for each Big Five trait.
\textbf{Combination-Level}: Individual method-personality pair stability scores.

\begin{table}[!h]
\centering
\small
\begin{tabular}{l|l|l|l}
\toprule
\textbf{Level} & \textbf{Category} & \textbf{Stability Score} & \textbf{Ranking} \\
\midrule
\multirow{3}{*}{Method} & ICL & 0.0366 & 1 \\
& PEFT & 0.0363 & 2 \\
& Steering & 0.0326 & 3 \\
\midrule
\multirow{3}{*}{Personality} & Openness & 0.0411 & 1 \\
& Conscientiousness & 0.0390 & 2 \\
& Extraversion & 0.0345 & 3 \\
\midrule
\multirow{3}{*}{Combination} & Steering+Conscientiousness & 0.0525 & 1 \\
& PEFT+Openness & 0.0456 & 2 \\
& ICL+Openness & 0.0407 & 3 \\
\bottomrule
\end{tabular}
\vspace{0.3em}
\caption{Top stability performers at each analysis level. Higher scores indicate better performance consistency across benchmarks.}
\vspace{-0.8em}
\label{tab:stability-rankings}
\end{table}
\vspace{0.2em}

\subsection{Limitations}

The stability metric oversimplifies complex performance trade-offs and focuses on academic benchmarks (MMLU, GAIA, BBQ). Normalization factors are empirically derived and may require adjustment for different model architectures.